\documentclass[accepted]{uai2023} 
\usepackage[american]{babel}

\usepackage{natbib} 
\usepackage{mathtools} 
\usepackage{booktabs} 
\usepackage{tikz} 

\usepackage{subfigure}
\usepackage{algorithm}

\usepackage{amsmath,amsfonts,bm}

\newcommand{\tblref}[1]{Table~\ref{#1}}

\def\Figref#1{Figure~\ref{#1}}

\def\Secref#1{Section~\ref{#1}}

\def\eqref#1{eq. ~\ref{#1}}
\def\Eqref#1{Eq.~\ref{#1}}

\def\Algref#1{Algorithm~\ref{#1}}

\def\1{\bm{1}}

\newcommand{\bld}[1]{\boldsymbol{#1}}

\def\rvf{{\mathbf{f}}}

\def\rvu{{\mathbf{i}}}

\def\rvk{{\mathbf{k}}}

\def\rvu{{\mathbf{u}}}

\def\rvx{{\mathbf{x}}}
\def\rvy{{\mathbf{y}}}

\def\rmK{{\mathbf{K}}}

\def\rmX{{\mathbf{X}}}
\def\rmY{{\mathbf{Y}}}
\def\rmZ{{\mathbf{Z}}}

\DeclareMathAlphabet{\mathsfit}{\encodingdefault}{\sfdefault}{m}{sl}
\SetMathAlphabet{\mathsfit}{bold}{\encodingdefault}{\sfdefault}{bx}{n}

\def\gB{{\mathcal{B}}}

\def\gD{{\mathcal{D}}}

\def\gF{{\mathcal{F}}}

\def\gL{{\mathcal{L}}}

\def\gN{{\mathcal{N}}}

\def\sR{{\mathbb{R}}}

\newcommand{\E}{\mathbb{E}}

\title{Guided Deep Kernel Learning}

\author[1]{\href{mailto:<Idan.Achituve@biu.ac.il>?Subject=Your UAI 2023 paper}{Idan Achituve}}
\author[2, 3]{Gal Chechik}
\author[1]{Ethan Fetaya}
\affil[1]{%
    Faculty of Engineering\\
    Bar-Ilan University\\
    Israel
}
\affil[2]{%
    Computer Science Dept.\\
    Bar-Ilan University\\
    Israel
}
\affil[3]{%
    NVIDIA\\
    Israel
  }
  
\begin{document}
\maketitle
\begin{abstract}
Combining Gaussian processes with the expressive power of deep neural networks is commonly done nowadays through deep kernel learning (DKL). Unfortunately, due to the kernel optimization process, this often results in losing their Bayesian benefits.
In this study, we present a novel approach for learning deep kernels by utilizing infinite-width neural networks. We propose to use the Neural Network Gaussian Process (NNGP) model as a guide to the DKL model in the optimization process. Our approach harnesses the reliable uncertainty estimation of the NNGPs to adapt the DKL target confidence when it encounters novel data points. As a result, we get the best of both worlds, we leverage the Bayesian behavior of the NNGP, namely its robustness to overfitting, and accurate uncertainty estimation, while maintaining the generalization abilities, scalability, and flexibility of deep kernels. Empirically, we show on multiple benchmark datasets of varying sizes and dimensionality, that our method is robust to overfitting, has good predictive performance, and provides reliable uncertainty estimations.


\end{abstract}

\section{Introduction}
Gaussian processes (GPs) are an effective Bayesian non-parametric family of models. They have several appealing features, such as tractable inference, accurate uncertainty estimation, and the ability to generalize well from small datasets \citep{gp_book, snell2020bayesian, achituve2021personalized}. In GPs, the kernel function is the crucial factor that determines their performance, as it measures the similarity between data points and significantly impacts which functions the model considers probable. Standard kernels, such as RBF kernels, perform well on certain learning problems, but they are inadequate for complex data modalities, like images and texts, failing to capture the desired semantic similarity. One appealing solution is to combine GPs with the expressive power of Neural Networks (NNs). There are two popular ways to achieve that. The first is through learning deep kernels, and the second is through kernels that correspond to infinite-width networks. In what follows we present both approaches, their limitations, and our proposed approach that combines the two. 


One popular way to combine GPs and NNs is through \textit{deep kernel learning} (DKL) \citep{calandra2016manifold, gordon16_DKL}. DKL uses a standard kernel over an embedding learned by a neural network, combining the tractable inference of GPs with the expressive power of deep neural networks (DNNs). Unfortunately, despite appearing to be a natural way to combine the benefits of GPs and DNNs, DKL often falls short of expectations in practice. A recent study found that deep kernels can severely overfit, sometimes even worse than standard NNs \citep{ober2021promises}. This work suggests that the DKL overfitting is caused by the optimization process  ``over-correlating'' the data points.



An alternative way to link DNNs and GPs, without relying on DKL, is through the equivalence between GPs and \textit{infinite-width} deep neural networks  \citep{neal2012bayesian, LeeBNSPS18, MatthewsHRTG18,Garriga-AlonsoR19, NovakXBLYHAPS19,Yang2019scaling}.  
Specifically, consider the distribution over DNN weights when they are initialized i.i.d. As the width of the DNN layers increases to infinity, the distribution of functions represented by the NN converges to a Gaussian process. Importantly, that GP has a kernel function that can be computed efficiently despite having an infinite width.  The main advantage of this approach is clear - it allows us to apply tractable Bayesian inference to highly expressive neural networks of infinite width. And, as the structure of DNNs provide valuable inductive biases for many data modalities, they can generate a corresponding kernel that is better suited to various data modalities. 

This approach, however, also has several drawbacks that hinder its widespread adoption.
 First and foremost, in many cases, these models underperform  standard NNs that were optimized for a specific task \citep{NovakXBLYHAPS19}. One possible explanation for this is that the success of DNNs is connected to the implicit bias in the optimization process (e.g. \citep{VardiS21}), which can not be captured by them. Second, the evaluation of the kernel in training and inference time can be costly. This is partially due to the fact that the NN kernel needs to be computed for every pair of data points, in comparison to DKL where we run the network on each datum once before applying a standard kernel. Finally, it is challenging to incorporate established mechanisms such as inducing point techniques with these types of models. This is in contrast to DKLs which are more flexible and easier to scale.
%
%
The limitations of current solutions raise the question: \emph{{How can we combine GPs with NNs without compromising performance or uncertainty estimation?}} 

This paper proposes a solution to the above question, which we call \textit{Guided Deep Kernel Learning} (GDKL). GDKL combines the benefits of DKL with NNGPs, by leveraging the uncertainty estimation of NNGPs to guide the DKL optimization process. To this end, we propose a novel procedure to optimize deep kernels by having them match the distribution of the NNGP's latent function given the target value. For example, consider a regression task, the DKL will try to match a Gaussian centered near the target with an adaptive level of certainty that depends on the NNGP. We show that this approach achieves the best of both worlds. It enjoys the flexibility, scalability, and predictive capabilities of DKL, while retaining the Bayesian benefits of GPs. Namely, our method can estimate uncertainty more reliably and is drastically less prone to overfitting than DKL, without sacrificing performance.  The experiments show the superiority of our method against natural baseline methods on several benchmark datasets in terms of both performance and uncertainty quantification.

This paper makes the following novel contributions. (i) We propose GDKL, a novel method to train deep kernels having their uncertainty calibrated by infinite-width networks; 
(ii) GDKL allows to perform either exact inference or approximate inference using common inducing point techniques; (iii) we demonstrate the benefits of GDKL over baseline methods for small- to mid-sized benchmark datasets with low and high data dimensionality. We conclude that GDKL generalizes well, can estimate uncertainty more reliably, and is significantly more robust against overfitting compared to standard deep kernels and competing methods.

\section{Background}
\paragraph{Notations.}
We denote scalars with lower-case letters (e.g., $x$), vectors with bold lower-case letters, (e.g., $\rvx$), and matrices with bold capital letters (e.g., $\rmX$). Given a dataset $\gD = \{(\rvx_1,\rvy_1),...,(\rvx_n,\rvy_n)\}$, we denote by $\rmX\in\sR^{n\times d}$ and $\rmY\in\sR^{n\times c}$ the design and label matrices whose $i^{th}$ row is $\rvx_i$ and $\rvy_i$ respectively.

\paragraph{Gaussian Processes.}
Gaussian processes are a family of Bayesian non-parametric models. GPs assume that the mapping from input points to the target values is via latent functions $\gF = \{f^1, ..., f^c\}$. In this study, we assume independence between the latent function values. Consider a single output dimension process $f(\cdot)$, a GP is fully specified by the mean function $m(\rvx)$ and the covariance function $k(\rvx,\rvx')$. We denote it by  $f(\rvx)\sim\mathcal{GP}(m(\rvx),~k(\rvx,\rvx'))$. The mean $m(\rvx)$ is commonly taken to be the constant zero function, and the kernel $k(\rvx,\rvx')$ is a positive semi-definite function. The kernel defines the correlation between function values at different input locations. Thus, it is the main contributing factor in predicting on novel inputs. One of the major benefits of GPs is that in regression tasks with Gaussian noise, $p(y_i|f(\rvx_i))=\mathcal{N}(f(\rvx_i),\sigma^2_n)$, the inference has a closed-form Gaussian solution. Specifically, we have analytical expressions for the posterior $p(f_*|\rvx_*,\gD)$ and the marginal $p(y_*|\rvx_*,\gD)$ where $\gD$ is the training data and $\rvx_*$ is a test data point. The hyper-parameters of the GPs, which we will refer to as parameters in this study, are commonly optimized using the marginal likelihood. Here, we promote the use of the predictive distribution to learn them. Several studies considered this approach in the literature (e.g., \citep{jankowiak2020parametric, snell2020bayesian, achituve2021personalized, lotfi2022bayesian}). Usually, this objective leads to better predictive abilities, yet as we will show, it is not robust against overfitting when training deep kernels.

\paragraph{Deep Kernel Learning}
In \citep{gordon16_DKL}, the authors proposed to combine deep neural networks with GPs by applying a GP on the representation learned by a NN. For example, consider the RBF kernel $k(\rvx,\rvx')=\exp(-||\rvx-\rvx'||^2/2\ell^2)$ (although any other kernel can be used),  they proposed the following kernel $k_\theta(\rvx,\rvx')=\exp(-||g_\theta(\rvx)-g_\theta(\rvx')||^2/2\ell^2)$ where $g_\theta$ is a NN with parameters $\theta$. They then trained $\theta$ to maximize the log marginal likelihood $\log(p(\rvy|\rmX))$ using the closed-form expression for regression problems. Later works extended this approach to classification \citep{linderman2015dependent, wilson2016stochastic, milios2018dirichlet, achituve2021gp}. 

\begin{figure*}[!t]
    \centering
    \includegraphics[width=0.40\textwidth]{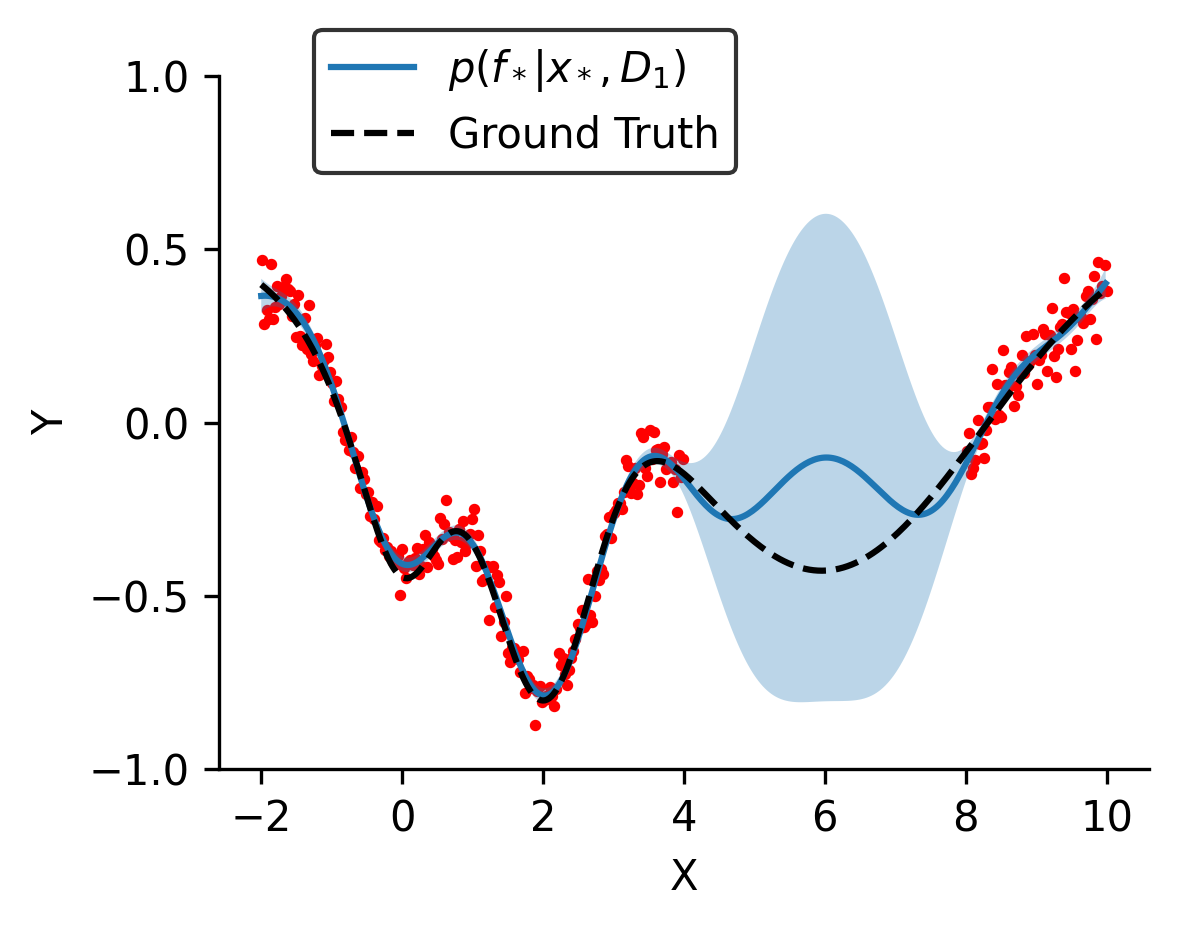}
    \includegraphics[width=0.40\textwidth]{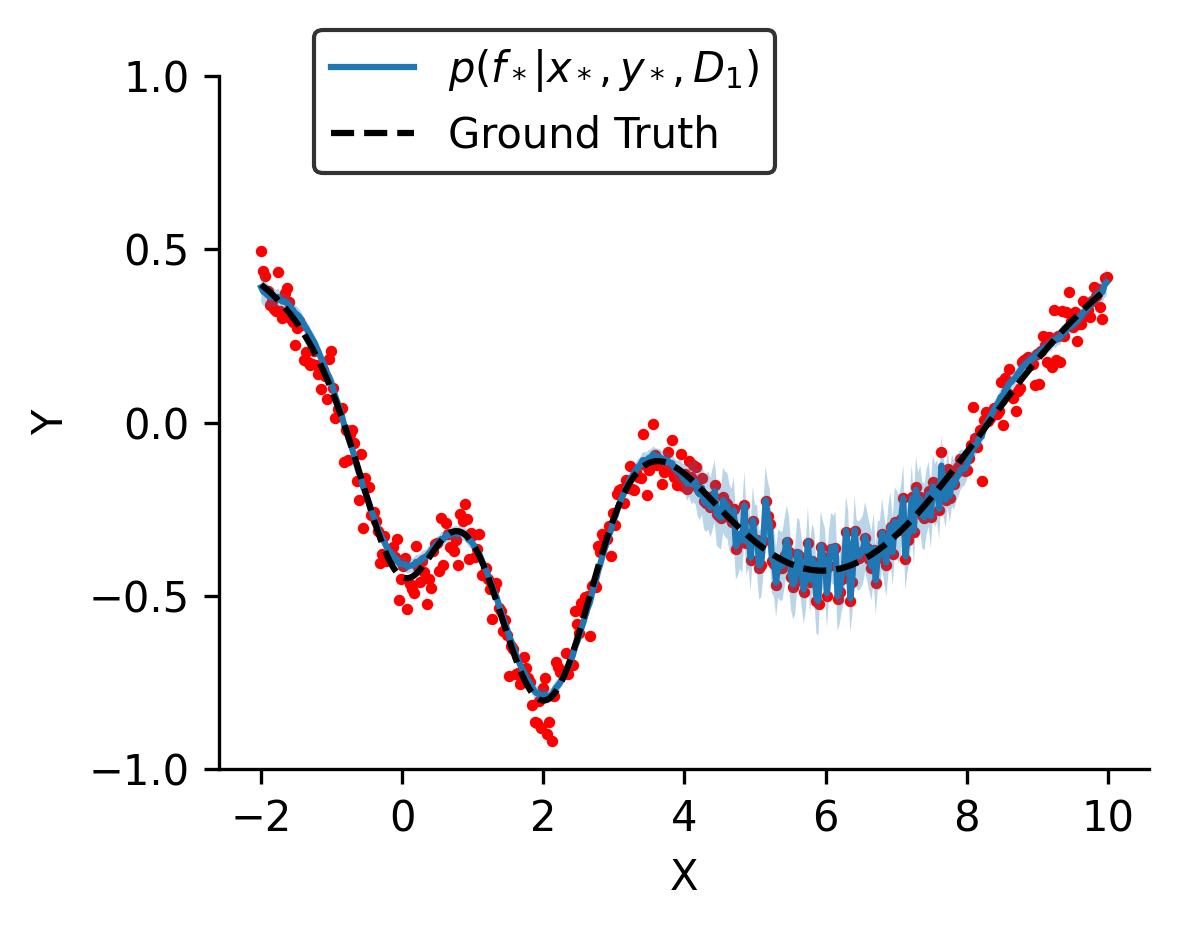}
\caption{Illustrative example: (Left) Points in $\gD_1$, the target function, and the GP prediction. Data contains a gap in [4,8] to demonstrate a low-confidence region.
(Right) Points in $\gD_2$ and the Gaussian objective $p(f_{*} | x_*, y_*, \gD_1)$ for each point.}
\label{fig:ill_example}
\end{figure*}

\paragraph{Infinite width networks.}
Studying the behavior of NNs in the infinite-width limit has its roots in the seminal work of \citet{neal2012bayesian}. It was shown that at initialization (with proper tuning), the distribution over functions represented by a single hidden layer NN converges to a GP as the width increases to infinity. This approach was later extended to infinite deep NNs as well \citep{LeeBNSPS18, MatthewsHRTG18}. This means that at the infinite-width limit, the Bayesian neural network inference problem is reduced to GP inference with a kernel defined by the neural network limit. In this study, we will refer to instances of this approach as the Neural Network Gaussian Process (NNGP). The kernel for a fully-connected network can be computed using the following recursive formula:
\begin{equation}
    \begin{aligned} 
    k^{(1)}(\rvx, \rvx') &= \sigma_b^2 + \sigma_w^2 \cdot \frac{\rvx^T \rvx'}{d}\\
    k^{(l+1)}(\rvx, \rvx') &= \sigma_b^2 + \sigma_w^2 \E_{f \sim \gN(0, \rmK^{(l)})}[ \phi(f(\rvx)) \phi(f(\rvx'))],
    \end{aligned} 
    \label{eq:NNGP}
\end{equation}
where $\sigma_b^2, \sigma_w^2$ are hyper-parameters which control the variances of the biases and weights respectively, $d$ and $l$ are the input dimension and layer index respectively (e.g., $\rmK^{(l)}$ denotes the kernel of the $l^{th}$ layer), and $\phi(\cdot)$ is the layer point-wise non-linear function. For some non-linear activations, such as sigmoidal, Gaussian, and Relu, the formula can be computed analytically \citep{williams1996computing, cho2009kernel}. In other cases, it can be approximated efficiently using Monte-Carlo methods as the expectation is over a two-dimensional Gaussian random variable \citep{NovakXHLASS20}. Similarly, a kernel can be derived for other NN architectures, such as CNNs and RNNs \citep{Garriga-AlonsoR19, NovakXBLYHAPS19, Yang2019scaling}.

\paragraph{Inducing points.}
One prominent limitation of GPs is the difficulty of doing exact inference on large datasets. Assuming we have a dataset with $n$ points, exact inference requires storing and inverting an $n\times n$ matrix. This operation imposes a memory and run-time complexity of $\Omega(n^2)$. A commonly used method to improve the scalability of GPs is through the use of inducing points (e.g., \citep{ titsias2009variational}). Inducing point methods define a set of pseudo-observations of size $m \ll n$, termed inducing locations. These locations may or may not be learned as part of the optimization process. Importantly, this mechanism allows us to control the size of the matrix to invert since in order to make predictions we can make all the costly operations only on these points instead of the actual dataset. 

\section{Method}
We now present and explain our method. We will first describe our approach in the setting of exact GP inference (i.e., without inducing points approximation), then we will show how our framework can be easily generalized to include inducing points, i.e the sparse GP case. Incorporating inducing points into our framework allows our approach to handle a wide range of problems, from limited-sized datasets, where the overfitting of DKLs is most severe, to large-scale problems where the NNGPs are too computationally demanding to run.  It is important to stress that we do not place any constraints on the NN architecture, unlike other existing solutions  \citep{liu2021deep, mallick2021deep, ober2021promises, van2021feature}. For the sake of clarity, we will describe our method for the case of a single output. The generalization to the multi-outputs case is immediate and will be discussed afterward.


\subsection{Guided Deep Kernel Learning}
\label{sec:GDKL}
Assume we are given a dataset $\mathcal{D}$. We split it into a training set $\mathcal{D}_1$ and a validation set $\mathcal{D}_2$. Denote by $p$ the NNGP model defined by an infinite-width neural network, and by $q_\theta$ the DKL model with parameters $\theta$. Given a point $\rvx_*$ we denote by $f_*$ the value of the latent GP function on $\rvx_*$, and denote by $D_{KL}$ the Kullback–Leibler (KL) divergence.

To motivate our proposed approach, we first describe two possible objectives, a Bayesian distillation objective (e.g., \citep{pensofunctional}) , and a predictive objective. Then, we present our final objective, which can be viewed as a combination of the two.


In distillation, we wish to train $q_\theta$ to mimic $p$. A natural way to achieve this goal is with the following objective:
\begin{equation}
    \label{eq:dist_obj}
    \ell_{dist}(\theta)=\E_{\rvx_*\sim\gD_2}D_{KL}[q_\theta(f_{*} | \rvx_*, \gD_1) || p(f_{*} |\rvx_*, \gD_1)].
\end{equation}
The objective in \Eqref{eq:dist_obj} tries to match the latent distribution of the two models on an unseen data point from $\gD_2$. Training $q_\theta$ may produce a model that behaves like the Bayesian NNGP model, but unfortunately, it will also inherit the subpar predictive performance of the NNGP model.

Alternatively, we can try to optimize the predictive distribution of the DKL model: 
\begin{equation}
\label{eq:pred_obj}
    \ell_{pred}(\theta)=\E_{(\rvx_*, y_*)\sim\gD_2}[-\log~q_\theta(y_*| \rvx_*, \gD_1)].
\end{equation}
This objective tends to produce accurate predictions as it is directly optimized to predict  $y_*$. However, it will not behave like a Bayesian model. Namely, it will suffer from the same overfitting issues as the standard DKL training does \citep{lotfi2022bayesian} which will result in an overestimated confidence. Appendix \ref{app_sec:obj_fun_analysis} empirically demonstrates this claim and the previous one. We evaluated both loss terms on the UCI datasets Boston, Concrete, and Energy, and found that indeed these losses behave as we anticipated. 

A possible middle ground between these two approaches is to optimize $q_\theta(y_*| \gD_1,\rvx_*)$  to match the distribution of $p(f_{*} |\gD_1, \rvx_*,y_*)$. The key difference is that in this case the latent variable $f_*$ is also conditioned on the sample $y_*$: 
\begin{equation}
\label{eq:hyb_obj}
    \E_{(\rvx_*, y_*)\sim\gD_2}D_{KL}[q_\theta(f_{*} | \rvx_*, \gD_1) || p(f_{*} | \rvx_*, y_*, \gD_1)].
\end{equation}
As the target latent distribution is conditioned on $y_*$, it will, for regression, take the form of a Gaussian centered near it, and the variance will be dependent on how confident $p(f_{*} |\rvx_*, y_*, \gD_1)$ is. In \Figref{fig:ill_example} we illustrate the usefulness of this objective on a toy problem. On the left panel we show the points in $\gD_1$, the ground truth function, as well as the posterior $p(f_{*} | x_*, \gD_1)$. We intentionally omitted points in the $[4,8]$ domain from $\gD_1$ to highlight areas where the GP is not confident. 
While the GP prediction in that area is not accurate, it appropriately assigns high uncertainty to its prediction. On the right panel, we show the points in $\gD_2$ and for each point, we show the target objective $p(f_{*} |x_*,y_*, \gD_1)$. We highlight two desired properties of our objective seen from this plot: When the GP is confident, the GP prediction is tight around the ground truth, and not centered around the noisy $y_*$ samples. However, when the GP is not confident $p(f_{*} |x_*,y_*, \gD_1)$ is centered around the noisy $y_*$ samples with a much larger variance. 

In Appendix \ref{app_sec:gdkl_objective} we show that the objective in \Eqref{eq:hyb_obj} is also equivalent to the following:
\begin{equation}
    \begin{aligned} 
     &\E_{(\rvx_*, y_*)\sim\gD_2} \E_{q_\theta(f_*| \rvx_*, \gD_1)}[-\log~p(y_* | f_*)] \\ &+D_{KL}[q_\theta(f_*| \rvx_*, \gD_1) || p(f_*|\rvx_*, \gD_1)].
    \end{aligned} 
    \label{eq:pred_with_reg}
\end{equation}
This representation makes it clear that our objective is in fact a combination of \Eqref{eq:dist_obj} and \Eqref{eq:pred_obj}. Specifically, \Eqref{eq:pred_obj} can be connected to the first term in \Eqref{eq:pred_with_reg} by marginalizing over $f_*$ and using Jensen inequality. Conceptually, one can think of \Eqref{eq:pred_with_reg} as having a data term, which is comparable to the DKL marginal likelihood objective, plus a regularizer that prevents the model from over-fitting the training points.

To gain further insight into our approach, consider estimating $\log~p(y_*|\rvx_*, \gD_1)$ using variational inference. While it has an analytical solution for a Gaussian likelihood, we can set it aside and derive the following evidence lower bound (ELBO):
\begin{equation}
    \small
    \begin{aligned} 
    &\E_{(\rvx_*, y_*)\sim\gD_2} \log~p(y_* | \rvx_*, \gD_1) = \\
    &\E_{(\rvx_*, y_*)\sim\gD_2} \log~\int \frac{q_{\theta}(f_*|\rvx_*, \gD_1)}{q_{\theta}(f_*|\rvx_*, \gD_1)}p(y_* | f_*)p(f_*| \rvx_*, \gD_1)df_* \geq \\
     &\E_{(\rvx_*, y_*)\sim\gD_2} \int q_{\theta}(f_*|\rvx_*, \gD_1)\log~\frac{p(y_* | f_*)p(f_*| \rvx_*, \gD_1)}{q_{\theta}(f_*|\rvx_*, \gD_1)}df_*.
    \end{aligned} 
    \label{eq:cmll}
\end{equation}
It is not hard to see that our objective is equivalent to maximizing the ELBO. Namely, $q_\theta$ is essentially trained as a variational distribution, similar to an encoder network in variational auto-encoders (VAEs) \citep{KingmaW13}. It thus tries to ``encode'' the label, but does not directly predict it. 

\begin{algorithm}[!t] 
    \caption{ Guided Deep Kernel Learning (\textit{GDKL})}
    \vspace{0.1cm}
    {\bf Input}: $\gD = (\rmX, \rvy)$ - the dataset; $\rmK$ - a pre-computed kernel of the NNGP on $\gD$; $T$ - number of training iterations; $\beta$ - a hyper-parameter that scales the KL-divergence term.\\
    {\bf Init} $\theta$, the parameters of the DKL.\\ 
    {\bf For} $i = 1, ..., T$:\\
    \hspace*{3mm} $\bullet$ Randomly split $\gD$ to $\gD_1$ and $\gD_2$, s.t. $\gD = \gD_1 \cup \gD_2$ \\
    \hspace*{3mm} and $\gD_1 \cap \gD_2 = \{\emptyset\}$\\
    \hspace*{3mm} $\bullet$ Construct $\rmK_{\gD_1}$ from $\rmK$ by selecting the \\   
    \hspace*{3mm} entries of examples from $\gD_1$\\
    \hspace*{3mm} \textbf{For} all $j \in \gD_2$: \\
    \hspace*{5mm} \textcolor{teal}{\# Exact expressions in the Appendix.}\\
    \hspace*{5mm} $\bullet$ Obtain the predictive posteriors $p(\rvf_j | \rvx_j, \gD_1)$, \\
    \hspace*{5mm} and $q_{\theta} (\rvf_j | \rvx_j, \gD_1)$ \\
    \hspace*{5mm} $\bullet$  $\gL^{D_{KL}}_j \leftarrow$ $  D_{KL}[q_\theta(f_j| \rvx_j, \gD_1) || p(f_j|\rvx_j, \gD_1)]$ \\
    \hspace*{5mm} $\bullet$ $\gL^{ELL}_j \leftarrow$ $  \E_{q_\theta(f_j| \rvx_j, \gD_1)}[-\log~p(y_j | f_j)]$ \\
    \hspace*{3mm} {\bf End for}\\
    \hspace*{3mm} $\bullet$ $\gL \leftarrow \frac{1}{|\gD_2|} \Sigma_{j=1}^{|\gD_2|}  \gL^{ELL}_j + \beta\cdot\gL^{D_{KL}}_j$. \\
    \hspace*{3mm} $\bullet$  Compute $\nabla_{\theta} \gL$ and perform update step.\\
    {\bf End for}
    \label{algo:GDKL}
\end{algorithm}

We now introduce two modifications to our objective. We add a  hyperparameter $\beta$ that multiplies the $D_{KL}$ term in  \Eqref{eq:pred_with_reg}. This allows us to control the balance between predictive training ($\beta=0$) and distillation ($\beta\rightarrow\infty$). Second, we utilize the fact that GPs are non-parametric and perform a random split of $\gD$ to different $\gD_1$ and $\gD_2$ sets at each iteration. We found that this approach led to better generalization compared to fixing these datasets. Hence the final objective is the following:

\begin{equation}
    \begin{aligned} L(\theta)=&\E_{\gD_1,\gD_2}\E_{(\rvx_*, y_*)\sim\gD_2}\{ \E_{q_\theta(f_*| \rvx_*, \gD_1)}[-\log~p(y_* | f_*)] \\ &+ \beta \cdot D_{KL}[q_\theta(f_*| \rvx_*, \gD_1) || p(f_*|\rvx_*, \gD_1)]\}.
    \end{aligned} 
    \label{eq:final_objective}
\end{equation}

Note that for the regression case, the $D_{KL}$ term and the expected log-likelihood term have a closed-form solution. In classification tasks, we can use approximations that involve a Gaussian likelihood, such as treating the classification problem as a regression problem, using transformed Dirichlet variables \citep{milios2018dirichlet}, or using the Pólya-Gamma augmentation \citep{polya_gamma, achituve2021gp}. In this study, we used the transformed Dirichlet variables technique. GDKL training procedure is illustrated in \Algref{algo:GDKL}. 



Finally, we would like to note a few technical details. 
First, we can evaluate the kernel of the NNGP once and extract at each iteration only the relevant sub-matrices. Second, the extension to multi-output GPs results in additional summation over each output dimension. Third, our network and the NNGP network do not have to share the same architecture (besides the obvious difference in width) and we have complete freedom in choosing the architecture. Lastly, to make predictions on novel data points we use the full dataset $\gD$ using the standard GP formulas. Namely, when making predictions our model is as fast as standard DKL models.

\begin{figure*}[!t]
    \centering
    \includegraphics[width=0.90\textwidth]{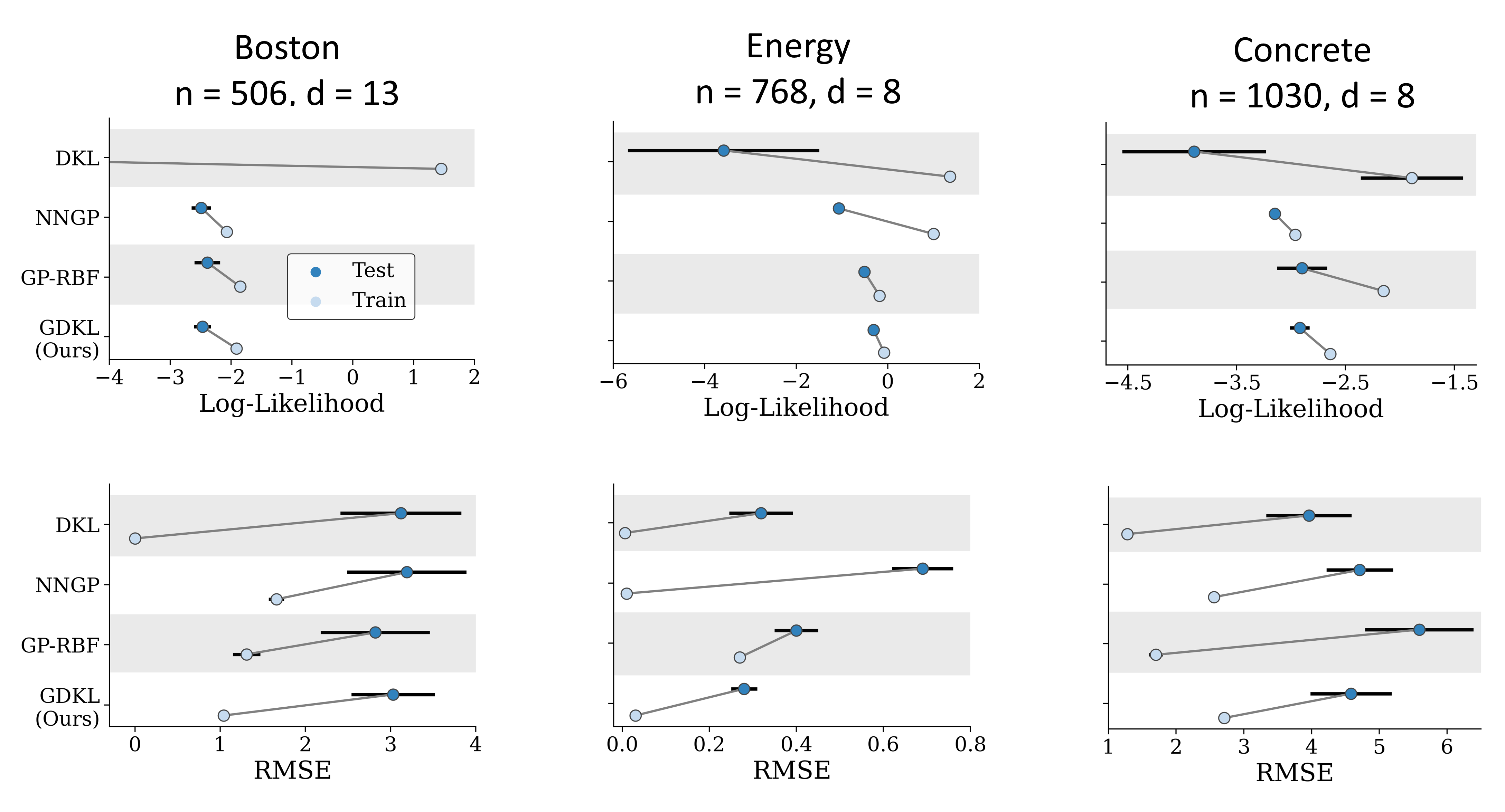}
\caption{Results for small UCI datasets. We report the log-likelihood (top; right is better) and RMSE (bottom; left is better) for each method over 10 splits on the training test sets. The log-likelihood of DKL on Boston is $\sim$-550.}
\label{fig:uci_small}
\end{figure*}

\subsection{Guided Deep Kernel Learning with Inducing Points}
Although the overfitting problem of DKLs is more acute in cases with limited data, it can still return overconfident predictions on large datasets. As such, we wish to extend our approach to larger datasets by incorporating inducing points. 
Denote by $\rmZ$ the set of $m$ inducing locations, and by $\rvu = \rvf(\rmZ)$ the function evaluation at these locations (i.e., the inducing variables). We follow the common practice \citep{hensman2013gaussian}, and define the posterior now as $q_{\theta}(\rvf) = \int p_{\theta}(\rvf|\rvu)q(\rvu)d\rvu$, where $p_{\theta}(\rvf|\rvu)$ is a Gaussian density according to the GP prior of the DKL, and $q(\rvu)$ is a variational Gaussian distribution with learned parameters. We note that while we omit $\rmZ$ for brevity, it plays an important role as the kernel matrix depends on $\rmZ$. Now we can plug this posterior distribution in \Eqref{eq:hyb_obj}:
\begin{equation}
    \begin{aligned} 
     \E_{\gD_1,\gD_2}\E_{(\rvx_*, y_*) \sim \gD_2} D_{KL} [q_{\theta}(f_* | \rvx_*, \rmZ) || p(f_* | \rvx_*, y_*, \gD_1)],
    \end{aligned} 
    \label{eq:hyb_obj_sparse}
\end{equation}

and obtain the objective in \Eqref{eq:final_objective} with the new posterior: 
\begin{equation}
    \begin{aligned} 
     &\E_{\gD_1,\gD_2}\E_{(\rvx_*, y_*) \sim \gD_2} \{ \E_{q_\theta(f_*| \rvx_*, \rmZ)}[-\log~p(y_* | f_*)] \\ & + \beta \cdot D_{KL}[q_\theta(f_*| \rvx_*, \rmZ) || p(f_*|\rvx_*, \gD_1)]\}.
    \end{aligned} 
    \label{eq:final_objective_sparse}
\end{equation}

A key part of scaling our objective is to allow for mini-batching. The objective in \Eqref{eq:final_objective_sparse} naturally factorizes over the data points in $\gD_2$, which leaves the $\gD_1$ terms. As we split our dataset at each iteration into a train and validation set, a simple solution is to split a random batch instead of the entire dataset. Namely, given a batch of examples $\gB$ we split it to two subsets $\gB_1$ and $\gB_2$ similar to the split we did for $\gD$ and compute the following objective:
\begin{equation}
    \begin{aligned} 
     &\E_{\gB_1,\gB_2}\E_{(\rvx_*, y_*)\sim\gB_2} \{ \E_{q_\theta(f_*| \rvx_*, \rmZ)}[-\log~p(y_* | f_*)] \\ & + \beta \cdot D_{KL}[q_\theta(f_*| \rvx_*, \rmZ) || p(f_*|\rvx_*, \gB_1)]\}.
    \end{aligned} 
    \label{eq:final_objective_sparse_batch}
\end{equation}


In this case, the inducing locations $\rmZ$ and the variational  parameters of $q(\rvu)$ are also learned as part of the optimization process. Nevertheless, we achieve two important goals. First, the DKL model still tries to match the posterior of the label-informed NNGP model. Second, we need to evaluate the NNGP model only on the actual data points which are known in advance and can be computed beforehand. Thus, we are avoiding costly evaluations on the inducing inputs by this model. Furthermore, we can define the inducing inputs in the feature space of the NN which is much more beneficial in terms of optimization compared to the input space of the network \citep{bradshaw2017adversarial, achituve2021gp}. In Appendix \ref{app_sec:comp_cons} we provide additional computational aspects of our method. Specifically, we discuss the scaling limitations imposed by computing the NNGP kernel to train GDKL.
We argue that with GDKL this is not so much of an issue as we have a a large degree of freedom in choosing the NNGP architecture with it hurting too much the performance of GDKL. Furthermore, with smart pre-computations GDKL can be as fast as standard DKL during the training time of the NN, and not only when making predictions.

\section{Related Work}
\textbf{Bayesian NNs.} Bayesian NNs model the uncertainty over the true underline function by assuming a probability distribution over the network parameters \citep{minka2000bayesian}. Instead of solving an optimizing process for a single set of parameters, BNNs attempt to compute the Bayesian model average (BMA) \citep{wilson2020bayesian}. However, for modern NNs solving the BMA integral is computationally intractable and approximations must be used. Notable examples are the Laplace approximation \citep{mackay1992bayesianint, khan2019approximate, daxberger2021laplace}, MCMC-based methods \citep{neal2012bayesian, welling2011bayesian, chen2014stochastic, ZhangLZCW20}, and variational inference \citep{graves2011practical, blundell2015weight, kingma2015variational}. These approximations usually result in either degraded performance, specialized NN architectures, unreliable uncertainty estimation, or computational difficulties in terms of memory and time. One possible compelling alternative for making inference in parameter space is to do it in function spaces \citep{sunfunctional, WangRZZ19, rudner2021tractable,  ma2021functional}. However, these methods suffer from similar issues as standard BNNs do and may involve rough approximations. A different line of work considers the distribution over functions when using infinite-width layers in fully connected networks \citep{neal2012bayesian, LeeBNSPS18, MatthewsHRTG18, jacot2018neural}. This approach allows performing tractable Bayesian inference with NNs while avoiding the optimization difficulties associated with training them. Later,
this approach was extended to other architectures and layers, such as CNNs, RNNs, attention, and batch normalization \citep{Garriga-AlonsoR19, NovakXBLYHAPS19, Yang2019scaling}. However, these approaches suffer from several drawbacks, such as reduced generalization, and costly kernel evaluation. In recent years several studies (e.g., \citep{aitchison2021deep, ober2021variational, yang2021theory}) extended this idea to introduce more flexibility to the kernel and the ability learn representation. Yet, to date, these methods suffer from reduced performance compared to standard NNs on large datasets, and they are not well adjusted to different data modalities and complex architectures. Lastly, there exists some evidence \citep{brosse2020last, kristiadi2020being} that being Bayesian, even only on the last layer, can provide the desirable benefits of the Bayesian paradigm.

\begin{figure*}[!t]
    \centering
    \includegraphics[width=0.90\textwidth]{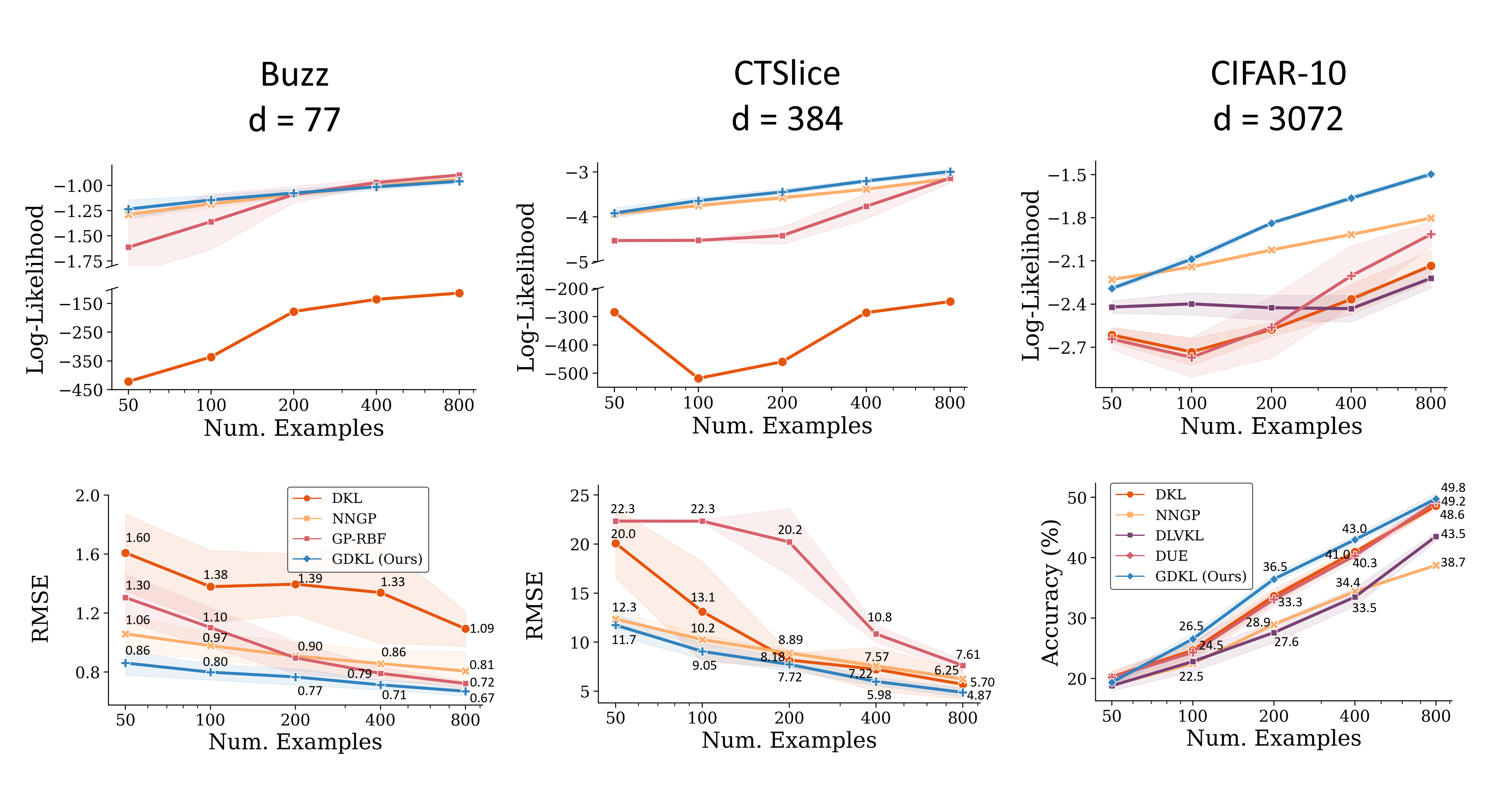}
    \caption{Model performance on $[50, 100, 200, 400, 800]$ training data points (x-axis in log-scale). We report the log-likelihood (top) and RMSE/accuracy (bottom) on the test set for all datasets. A higher log-likelihood and accuracy, and a lower RMSE are better. All the results are based on ten random seeds. On Buzz and CTSlice we didn't report here the standard deviation of the log-likelihood for the DKL model as it was very large and impaired the visibility of the figure.}
    \label{fig:high_dim_data}
\end{figure*}

\textbf{Learning representations with GPs.} 
An alternative approach for learning in function spaces is with Gaussian processes.
However, Gaussian processes cannot learn a new representation of the data \citep{gordon16_DKL}. Effectively this limits them to data modalities on which standard kernels can capture similarity well. Common solutions for this problem are deep GPs \citep{damianou2013deep, salimbeni2017doubly}, and deep kernel learning \citep{calandra2016manifold, gordon16_DKL}. In this study, we build on the latter approach.
Unfortunately, it was found that DKLs can overfit in a particular way \citep{ober2021promises}. The DKL objective will tend to correlate all data points instead of only those that convey information about each other. One way to mitigate this phenomenon is to use a fully Bayesian approach \citep{ober2021promises}. Yet, this direction inherits the challenges of working with BNNs which one would like to avoid when using DKLs. 
Simultaneously, several studies suggested methods to tackle this limitation of DKLs.  
\citet{van2021feature} proposed to do spectral normalization to the NNs parameters in architectures with residual connections following \citep{liu2020simple}. However, this method is limited to networks with residual connections only and may depend heavily on the estimation quality of the spectral norm. Also, in our experiments we often found this method to be equivalent to standard DKLs. \citet{liu2021deep} proposed to use stochastic NNs to learn the representations of examples. The first method, termed DLVKL, uses an encoder network, similar to that used in VAEs \citep{KingmaW13}. The second method, termed DLVKL-NSDE, uses stochastic differential equation flows. To use flow-based models, the feature and input spaces must have the same dimensionality which makes this method impractical for high-dimensional data.
Lastly, \citet{mallick2021deep} proposed to map data points to probability distributions using probabilistic NNs and fit a GP in that space. This method builds on particle-based optimization \citep{liu2016stein} and as such it operates on several NNs simultaneously which may be challenging for even moderate-sized NNs. We would like to note, that we expect our method to gain benefit from similar techniques (e.g., \citep{lakshminarayanan2017simple}). We leave this direction to future research endeavors.

\section{Experiments}
We evaluated GDKL on a number of benchmark regression and classification datasets, ranging from small to medium size, and low to high data dimensionality. Unless stated otherwise, in all experiments we report the mean performance (e.g., log-likelihood) along with one standard deviation over random seeds, which may include randomness in the data and the parameters. We stress that we used the same initialization for all compared methods. Note that in our evaluation we aim at showing that GDKL can obtain a strong mean prediction with good uncertainty estimation, while other methods usually fall short in at least one aspect. Indeed, as we will show, when factoring both (important) aspects GDKL is the best in most cases.
Full implementation details are given in Appendix \ref{app:exp_details}, comparison to baseline methods in terms of computational complexity are presented in Appendix \ref{app_sec:comp_cons}, and further experiments are presented in Appendix \ref{app_sec:add_exp}. 
\footnote{Our code is publicly available at \textcolor{magenta}{\url{https://github.com/IdanAchituve/GDKL}}}.

\begin{table*}[!h]
\centering
\caption{Test results on full CIFAR-10 and CIFAR-100 based on three random seeds.}
\scalebox{.73}{
    \begin{tabular}{l c cccccccc cccccccc}
    \toprule
    && \multicolumn{7}{c}{CIFAR-10} && \multicolumn{7}{c}{CIFAR-100}\\
    \cmidrule(l){3-9}  \cmidrule(l){11-17}
    && ACC ($\uparrow$) && LL ($\uparrow$) && ECE ($\downarrow$) && MCE ($\downarrow$) && ACC ($\uparrow$) && LL ($\uparrow$) && ECE ($\downarrow$) && MCE ($\downarrow$)\\
    \midrule
    DKL && 95.45 $\pm$ 0.06 && -0.19 $\pm$ 0.00 && 0.03 $\pm$ 0.00 && 0.30 $\pm$ 0.01 && 77.90 $\pm$ 0.45 && -0.94 $\pm$ 0.00 && 0.08 $\pm$ 0.00 && 0.24 $\pm$ 0.02 \\
    DLVKL && \textbf{95.65 $\pm$ 0.12} && -0.18 $\pm$ 0.00 && 0.03 $\pm$ 0.00 && 0.42 $\pm$ 0.20 && 77.42 $\pm$ 0.05 && -0.95 $\pm$ 0.02 && 0.09 $\pm$ 0.00 && 0.26 $\pm$ 0.02\\
    DUE && 95.48 $\pm$ 0.09 && -0.19 $\pm$ 0.00 && 0.03 $\pm$ 0.00 && 0.32 $\pm$ 0.04 && 76.39 $\pm$ 0.23 && -0.98 $\pm$ 0.03 && 0.10 $\pm$ 0.01 && \textbf{0.16 $\pm$ 0.02}\\
    \midrule
    GDKL (Ours) && \textbf{95.67 $\pm$ 0.06} && \textbf{-0.17 $\pm$ 0.00} && \textbf{0.01 $\pm$ 0.00} && \textbf{0.27 $\pm$ 0.01} && \textbf{78.36 $\pm$ 0.19} && \textbf{-0.89 $\pm$ 0.01} && \textbf{0.06 $\pm$ 0.00} && 0.18 $\pm$ 0.01\\
    \bottomrule
    \end{tabular}
}
\label{tab:full_datasets}
\end{table*}

\subsection{Small-Sized Datasets} \label{sec:small_sized_data}
To showcase our claim that GDKL can learn from small datasets while being robust to overfitting, we first evaluated it on the three small-sized UCI benchmark datasets: Boston, Energy, and Concrete \citep{asuncion2007uci}. We compared the exact GP variant of our method to (1) \textit{DKL} - standard DKL training \citep{calandra2016manifold, gordon16_DKL}; (2) \textit{NNGP} - A GP with an NNGP  kernel \citep{LeeBNSPS18, MatthewsHRTG18}; and (3) \textit{GP-RBF} - Standard GP with an RBF kernel (without DKL). The last simple baseline is considered a strong approach on these datasets \citep{salimbeni2017doubly}, as the semantic similarity is well captured by the RBF kernel. All neural network models (NNGP, DKL, and GDKL) use a three hidden layer fully-connected network. DKL and GDKL use the same width for each layer. We follow the training protocol suggested in \citep{ober2021promises} with several modifications which are described in Appendix \ref{app:exp_details}. As customary on these datasets (e.g., \citep{salimbeni2017doubly}), we use $k$-fold cross validation with $90\%$ randomly selected data as training and the remaining $10\%$ as a held-out test set. Here we used $k=10$. We scale the inputs and outputs of each partition of the data to have zero mean and unit standard deviation based on the training part only (the output scaling is restored in evaluation). \Figref{fig:uci_small} shows the Log-Likelihood (LL) and RMSE of the compared methods. 

From the figure, we observe several findings. First, we indeed observe that standard DKL training produces a model that has the characteristics of a NN and not a Bayesian model, i.e. it has good RMSE values at train and test, but it does not reliably estimate its uncertainty, as seen from its inferior test log-likelihood values. Second, the RMSE performance of NNGP on the Energy dataset is considerably worse than other baselines, confirming our claim that NNGP can have poor predictive performance.  
Finally, GDKL is always comparable to the best of the two on all datasets on both metrics and is substantially less prone to overfitting. 


\subsection{high-dimensional Datasets} \label{sec:high_dim_data}
Next, we expect to achieve the most benefit from DKL in settings with high dimensional data, where standard kernels do not perform as well. In these cases, the NN should be encouraged to find a low-dimensional representation over which a GP will work well. To test that scenario we considered the two regression datasets Buzz and CTSlice from the UCI repository, and the classification dataset CIFAR-10 \citep{krizhevsky2009learning}. Here, we use a subset of the training data with a  varied number of training examples from $50$ to $800$ and recorded the log-likelihood and the RMSE/accuracy of the model on the test set in each experiment. For Buzz and CTSlice we allocated $10\%$ of the data for testing, and for CIFAR-10 we use the default test split.
On the regression datasets, we compared the exact GP variant of our method to the same baselines as in \Secref{sec:small_sized_data}. On CIFAR-10, we use the Dirichlet-based likelihood function suggested in \citep{milios2018dirichlet} for inference. On this dataset, we didn't compare to the GP-RBF baseline as it works poorly on images. However, we did compare to two additional baselines: (1) \textit{DLVKL} \citep{liu2021deep} which learns a stochastic encoder network, reminiscent of VAEs \citep{KingmaW13}, to promote a regularized representation of the data; and (2) \textit{DUE} \citep{van2021feature} which applies spectral normalization on the weights with architectures that contain residual connections. Note that unlike GDKL these baselines require some modification to the NN. Here, we used a variant of the wide residual network (WRN) \citep{ZagoruykoK16} as a feature extractor in DKL models. As for the NNGP baseline (and for modeling $p$ in GDKL), we used a variant of this network without the average pooling layer as it imposes a large computational burden. The results on the three datasets are shown in \Figref{fig:high_dim_data}.

From the figures, we observe again that the DKL model overfits strongly, and in some cases, even its mean prediction is substantially lower than baseline methods. In addition, the NNGP works well on the regression datasets, but less so on real images. And finally, here as well, across all training set sizes, GDKL achieves the highest, or comparable, results in both the log-likelihood and RMSE/accuracy. In Appendix \ref{app_sec:reliability_diagrams} we also quantify the uncertainty of the models through calibration on the CIFAR-10 dataset. We compare all methods both visually using reliability diagrams and common metrics \citep{brier1950verification, guo2017calibration} on all dataset sizes. The figures show that GDKL is best calibrated across all metrics in all cases when $n \geq 200$, and on smaller dataset sizes it is second only to the NNGP model.

\subsection{Medium-Sized, high-dimensional Datasets}
\label{sec:large_dataset}
Having established that GDKL works well in low-data regime settings, we now evaluate its performance on larger datasets in which exact inference is more challenging. We do so on the full CIFAR-10 and CIFAR-100 datasets. We compare GDKL to the standard DKL baseline, and to DLVKL and DUE which were presented in \Secref{sec:high_dim_data}. In general, we followed the protocol suggested in \citep{van2021feature} for training on the CIFAR-10 dataset having only $10$ inducing points. For CIFAR-100 we used a similar protocol with $200$ inducing points. Exact experimental details are given in Appendix \ref{app:exp_details}. Here, as well, we used a variant of the WRN for both the DKL models and the NNGP model used by GDKL. \tblref{tab:full_datasets} shows the test accuracy, log-likelihood, expected calibration error (ECE), and maximum calibration error (MCE) for both datasets. The ECE measures a weighted average distance between the classifier's confidence and accuracy, and the MCE measures the maximum instead of the average. From the table, GDKL outperforms all baselines in almost all of the cases. Note how GDKL is able to maintain and even surpass the accuracy of DKL while providing a classifier that is better calibrated.

\section{Conclusions}
In this study, we put forward a novel method for learning deep kernels. Our goal is to train deep kernels that keep the benefits of Bayesian models without sacrificing performance. To this end, we define a new training procedure that uses an infinite-width NN to guide the DKL optimization, effectively setting adaptive levels of confidence in our predictions. This objective utilizes the reliable uncertainty estimation of NNGPs to allow our model to be as confident as possible without being over-confident. Finally, we also proposed an extension of our model to incorporate inducing points. We evaluated GDKL on small to mid-sized datasets having low and high data dimensionality. We found that our method consistently generalized well to novel data points while not scarifying the Bayesian properties of it, i.e., it doesn't overfit. As a possible future research direction, it would be interesting to combine our framework with Bayesian models other than infinite-width NNs. 

\section{Acknowledgements}
This study was funded by a grant to GC from the Israel Science Foundation (ISF 737/2018), and by an equipment grant to GC and Bar-Ilan University from the Israel Science Foundation (ISF 2332/18). IA is supported by a PhD fellowship from Bar-Ilan data science institute (BIU DSI). 


\bibliography{references}
\clearpage

\title{Guided Deep Kernel Learning - Supplementary Material}
  
\onecolumn 

\appendix
\section{Experimental Details}
\label{app:exp_details}
All experiments were done with GPyTorch \citep{gardner2018gpytorch} on NVIDIA GeForce RTX 2080 Ti having 11GB of memory. To compute the kernel of the NNGP we used the Neural Tangents library \citep{NovakXHLASS20}.

\textbf{Toy Dataset.} To construct the toy example we define the following target function (following  \citep{leclercq2018bayesian}): $f(x) = 0.6 - e^{-(x - 2)^2} - e^{-\frac{1}{10}(x - 6)^2} - \frac{1}{x^2+1}$. We sample uniformly at random $800$ points in $[-2, 12]$, evaluate the function on them and add observation noise of $\sigma_n^2 = 0.05$. Then we partition the dataset by random sampling to two subsets of $400$ points each, namely $\gD_1$ and $\gD_2$. We remove from $\gD_1$ all the points from the domain $[4, 8]$, and fit a GP with an RBF kernel to this dataset. We learn the hyper-parameters of this kernel using the ADAM optimizer \citep{KingmaB14} with the log marginal likelihood. Then, we evaluate $p(f_* | x_*, \gD_1)$ and $p(f_* | x_*, y_*, \gD_1)$ for all $(x_*, y_*) \in \gD_2$. 

\textbf{UCI.} We followed most of the training protocol suggested in \citep{ober2021promises}. 
To download and manipulate the datasets we used the Bayesian benchmarks git repository: {\href{https://github.com/hughsalimbeni/bayesian_benchmarks}{[https://github.com/hughsalimbeni/bayesian\_benchmarks]}}. 
To train the models, on Boston, Concrete, and Energy we perform 10-fold cross-validation using random seeds according to $90\%-10\%$ train-test splits. On Buzz and CTSlice we perform 3-fold cross-validation. We computed the normalization statistics (i.e., mean and std) based on the train split only and normalize all the data using them for model fitting. However, the results shown in the paper are on the unnormalized target values (i.e., the original values) which differ only in the scale. In all experiments, we used a fully connected network with the following architecture $[d, 100, 100, 100, 20]$ and ReLU activations. We used the same number of layers and activation for the infinite-width network. We initialized the variance of the observation noise to $\sim 0.02$ and learned it along with the model parameters. We used a weight decay of $1e-4$ for the DKL model only (no weight decay for GDKL), and we set the variance of the weights and biases of the NNGP to $1.6$ and $0.2$ as we found these values to work well on several toy examples. We trained all baseline models for $8000$ iterations. In GDKL we first pre-train the NNGP model observation noise and output scale of the kernel for $1000$ iterations, and then we train the DKL model for another $7000$ iterations in order to be comparable in the number of gradient steps used by the baseline methods. Also, on Buzz and CTSlice we used $\beta = 1.2$ as we found it to work slightly better than $1$ on a predefined validation set. We used a learning rate of $1e-2$ which drops by a factor of $10$ after $60\%$ and $80\%$ of the training (not including the pre-train stage of GDKL).


\textbf{CIFAR-10/100.} CIFAR-10 and CIFAR-100 \citep{krizhevsky2009learning} contain 60K images each with 10 and 100 distinct classes respectively. We used the default train-test split of 50K-10K. To perform a hyperparameters search, we allocated 5K examples from the training set. To report the results in \Secref{sec:large_dataset} in the main text, we use all of the training data (i.e., training set and validation set). In all experiments on these datasets we used Wide Residual Networks \citep{ZagoruykoK16} with a widen factor $k = 5$ so it will fit in the GPU. We used the features obtained in the last layer, after applying average pooling, as the input to the GP layer for DKL, DUE, and GDKL. As for the DLVKL baselines, we used an additional linear layer of size $100$ which we split into two halves for the mean and variance vectors of the Gaussian. For the NNGP model we used the same network, but without the average pooling layer as it imposed a large computational burden. We note that this step may harm the performance of the NNGP model \citep{NovakXBLYHAPS19}. On these datasets, for all DKL-based methods, we used a dropout rate of $0.3$.

On CIFAR-10 experiments in \Secref{sec:high_dim_data} we leverage the Dirichlet likelihood function suggested in \citep{milios2018dirichlet} with $\alpha_\epsilon = 0.01$. To make predictions with this likelihood one needs to sample from the posterior of $f_*$. Hence during test time, we sampled $1024$ values. During the training of GDKL, we sampled $256$ values as it uses the predictive distribution to train the model. We train all methods for a total of $7000$ gradient steps. For GDKL, we use the first $1000$ iteration to pre-train the NNGP model hyper-parameters and then train the DKL model for another $6000$ iterations. We used SGD with momentum of $0.9$, and an initial learning rate of $1e-2$ that drops by a factor of $10$ after $60\%$ and $80\%$ of the training (not including the pre-training stage for GDKL). We used a weight decay of $5e-4$ in all DKL-based methods except for GDKL which was set to $0$.
We did a grid search to select the best hyper-parameters for each method based on the validation set. The DLVKL objective has two KL divergences. We applied a grid search over their coefficients in $\{0.01, 0.1, 1.0\}$. Since DLVKL is based on variational inference, we set the number of inducing points to be the minimum value between the number of examples and $200$. The inducing locations were initialized using k-means. As in the official code of this baseline, we used a prior over the latent variable $z$ having a unit variance and a mean value that corresponds to the PCA projection of the input data. For the DUE baseline, we searched over the normalization coefficient and number of power iterations in $\{1, 3\}$, and for the NNGP model we searched over the weight variance in $\{1, 3, 5\}$ while keeping the variance of the bias fixed at $0.2$. We found that performance was similar for all values and picked the value $5$. For GDKL we also searched over the parameter $\beta \in \{0.1, 1.\}$ and we found that using $\beta=1.$ generated better results. Note that since this is a multi-output learning setup the KL divergence in GDKL objective results in a summation over the classes. To make it invariant to the number of classes we take an average instead of a sum. This is effectively the same as scaling the KL divergence term by another factor that equals to $0.1$. We did not perform data augmentations in these experiments.


In the experiments of \Secref{sec:large_dataset}, for the most part, we followed the protocol suggested in \citep{van2021feature}. Here, we used the Softmax likelihood function for training the deep kernels of all methods. To train the models we used $16$ samples from the latent GP when computing the likelihood, and on novel test points, we used $320$ samples. On CIFAR-10 we set the number of inducing points to $10$, and on CIFAR-100 to $200$ for all methods. All inducing locations were initialized using k-means. We set the number of training epochs to $200$ with a batch size of 256. To train GDKL we initially sampled $5\%$ of the data to train the hyper-parameters of the NNGP model for $1000$ iterations which correspond to $\sim 5$ epochs with a batch size of 256. Then we train the DKL model for an additional $194$ epochs. We used SGD with momentum of $0.9$, and an initial learning rate of $1e-1$ that drops by a factor of $10$ after $50\%$ and $75\%$ of the training (not including the pre-training stage for GDKL). We used a weight decay of $5e-4$ for all methods. Here we used $\beta=0.1$ for GDKL. Also, as in the exact setting of GDKL, we approximate the objective in Eq. 10 with MC samples. Note, however, that unlike Eq. 7 where $\gD_1$ appears in all terms of the objective, here the corresponding element $\gB_1$ appears only in the posterior distribution of the NNGP model.  Thus, to be more data efficient, we use two samples. One with $\gB_1$ as the "observed data" and another one with $\gB_2$ as the "observed data". In DUE and DKL, we also searched over the coefficient of the KL divergence in the variational ELBO objective in $\{0.1, 1.0\}$. We used random cropping and random horizontal flip for data augmentation.

\section{The GDKL Objective}
\label{app_sec:gdkl_objective}

In \Secref{sec:GDKL} we presented to following objective:
\begin{equation}
\label{app_eq:hyb_obj}
    \E_{\rvx_*,y_*\sim\gD_2} D_{KL} [q_\theta(f_{*} | \rvx_*, \gD_1) || p(f_{*} |y_*, \rvx_*, \gD_1)].
\end{equation}
We now show that it is equivalent to the objective of Eq. 6 in the main text.

\begin{equation}
\label{app_eq:hyb_obj_derivation}
\begin{aligned}
    &\E_{\rvx_*,y_*\sim\gD_2}D_{KL} [q_\theta(f_{*} | \rvx_*, \gD_1) || p(f_{*} |\rvx_*, y_*, \gD_1)] \\
    &=\E_{\rvx_*,y_*\sim\gD_2} \E_{  q_\theta(f_{*} | \rvx_*, \gD_1)} [\log~\frac{q_\theta(f_{*} | \rvx_*, \gD_1)}{p(f_{*} |\rvx_*, y_*, \gD_1)}] \\
    &= \E_{\rvx_*,y_*\sim\gD_2} \E_{  q_\theta(f_{*} | \rvx_*, \gD_1)}[\log~q_\theta(f_{*} | \rvx_*, \gD_1) - \log~\frac{p(y_* |f_{*}) p(f_{*}| \rvx_*, \gD_1)}{p(\rvx_*, \gD_1)}] \\
    &= \E_{\rvx_*,y_*\sim\gD_2} \E_{  q_\theta(f_{*} | \rvx_*, \gD_1)}[\log~q_\theta(f_{*} | \rvx_*, \gD_1) - \log~p(y_* |f_{*}) - \log~p(f_{*}| \rvx_*, \gD_1) + \log~p(\rvx_*, \gD_1)]\\
    &\propto \E_{\rvx_*,y_*\sim\gD_2} \E_{  q_\theta(f_{*} | \rvx_*, \gD_1)}[- \log~p(y_* |f_{*})] + D_{KL}[q_\theta(f_{*} | \rvx_*, \gD_1) || p(f_{*}| \rvx_*, \gD_1)].
\end{aligned}
\end{equation}
Where, in the third step we used Bayes rule, and in the last step we dropped the constant factor $\log~p(\rvx_*, \gD_1)$ which doesn't effect the optimization process.
\\~\\
For a Gaussian likelihood, the posterior predictive distributions can be derived using standard Gaussian algebra \citep{gp_book}. For instance,
\begin{equation}
    \begin{aligned} 
    & p(f_*|\rvx_*, \gD_1)=\mathcal{N}(\mu_*^p, (\sigma_*^p)^2),\\
    &\mu_*^p=\rvk_{*}^T(\rmK+\sigma^2_n \bld{I})^{-1}\rvy,\\
    &(\sigma_*^p)^2 = k_{**} - \rvk_*^T(\rmK+\sigma^2_n\bld{I})^{-1}\rvk_*.
    \end{aligned} 
\end{equation}
Where, $K_{ij}=k(\rvx_i,\rvx_j)$, $k_{**} = k(\rvx_*, \rvx_*)$, and $\rvk_*[i]=k(\rvx_i, \rvx_*)$.
In a similar fashion $q_{\theta}(f_{*}| \rvx_*, \gD_1) = \gN(f_{*} | \mu_*^q, (\sigma_*^q)^2)$ can be obtained.

Now, the $D_{KL}$ term has the following closed-form solution:
\begin{equation}
\label{app_eq:kl}
\begin{aligned}
D_{KL}[q_\theta(f_{*} | \rvx_*, \gD_1) || p(f_{*}| \rvx_*, \gD_1)] = \log~\frac{\sigma_*^p}{\sigma_*^q} + \frac{(\sigma_*^q)^2 + (\mu_*^q - \mu_*^p)^2}{2 (\sigma_*^p)^2} - \frac{1}{2}.
\end{aligned}
\end{equation}

Similarly, the expected log-likelihood term can be computed analytically using the following:
\begin{equation}
\label{app_eq:ell}
\begin{aligned}
\E_{  q_\theta(f_{*} | \rvx_*, \gD_1)}[- \log~p(y_* |f_{*})] = \frac{1}{2}(\log~2\pi + \log~\sigma_n^2 + \frac{(y_* - \mu_*^q)^2 + (\sigma_*^q)^2}{\sigma_n^2}).
\end{aligned}
\end{equation}

\section{Computational Considirations}
\label{app_sec:comp_cons}
In this section we address the computational complexity of GDKL from two aspects: (1) the scaling limitations imposed by the NNGP kernel, and (2) comparison to baseline methods.

\textbf{Scaling limitations imposed by the NNGP kernel.} GDKL leverages NNGP kernels to learn the model. One may wonder if that may pose a limit to GDKL as computing the NNGP kernel can be costly. To address this concern we provide two important computational aspects to showcase that it is not an issue for GDKL. Furthermore, we argue that GDKL posses computational advantages over using NNGPs directly, aside from the added benefit in performance.

\begin{itemize}
    \item First, as with standard NNGPs, GDKL inherits the flexibility in choosing the architecture of the NNGP and other design choices, such as the data resolution on which this kernel is computed. However, unlike traditional NNGPs where the posterior is heavily influenced by the NNGP kernel, in GDKL one may choose simpler kernels that possibly can be computed more efficiently without it resulting in a significant performance drop. This is because the NNGP basically serves as a prior in our model, aimed at calibrating the uncertainties of the DKL model. Hence, as we see it, GDKL can provide practitioners with the freedom to choose an NNGP model based on their hardware constraints, without compromising on model performance significantly. To validate that point, in \tblref{app_tab:shallower_NNGP} we present the effect of using only one residual block in each group, instead of four, in the kernel obtained by the wide-residual network architecture which we used throughout. The table shows the results in terms of test accuracy under the setup of \Secref{sec:high_dim_data}, but a similar trend was observed in terms of log-likelihood as well. The results in the table indicate a clear advantage, albeit small but still statistically significant, to the deeper NNGP model compared to the shallower one. However, when using the shallower NNGP model in the training process of GDKL, it has almost no effect on the test accuracy of GDKL.

    \begin{table*}[!h]
    \centering
    \caption{Effect of infinite network depth - test accuracy on CIFAR-10 with $\{50, 100, 200, 400, 800\}$ training examples.}
    \scalebox{.85}{
        \begin{tabular}{l c ccccc}
        \toprule
        && 50 & 100 & 200 & 400 & 800\\
        \midrule
        NNGP - One Res. Block &&  18.80 $\pm$ 0.11 & 21.10 $\pm$ 0.11 & 27.71 $\pm$ 0.13 & 32.18 $\pm$ 0.11 & 36.34 $\pm$ 0.09\\
        NNGP - Four Res. Blocks &&  18.87 $\pm$ 0.16 & 22.47 $\pm$ 0.16 & 28.94 $\pm$ 0.13 & 34.39 $\pm$ 0.09 & 38.73 $\pm$ 0.07\\
        \hline
        GDKL - One Res. Block && 19.41 $\pm$ 0.05 & 26.70 $\pm$ 1.11 & 36.52 $\pm$ 0.94 & 42.60 $\pm$ 0.53 & 49.34 $\pm$ 0.52\\
        GDKL - Four Res. Blocks && 19.35 $\pm$ 0.65 & 26.53 $\pm$ 0.98 & 36.45 $\pm$ 0.72 & 42.97 $\pm$ 0.71 & 49.75 $\pm$ 0.94\\
        \bottomrule
        \end{tabular}
    }
    \label{app_tab:shallower_NNGP}
    \end{table*}

    \item Second, in terms of computation of the kernel. When dealing with small to medium-sized datasets, computing the NNGP kernel is not typically expensive and can be done offline before training. However, for larger datasets, scalability becomes more challenging. In these cases, usually one will need to deal with scalability issues of GPs in general, and the common practice is to use inducing point (IP) methods such as the one we proposed in the paper.  
    In the IP variant of GDKL, batches are sampled during training and the NNGP kernel depends only on the examples in them. Therefore, one option is to compute the NNGP kernel online based on the examples in each batch. Since the batch size is usually small (e.g., 256), efficient optimization packages (e.g., \citep{NovakXHLASS20}) can be utilized to compute these kernel matrices in a relatively efficient manner. However, this approach may still be slower than training standard DKLs. Therefore, a further improvement can be done with proper engineering work. One can pre-compute the kernel of the examples in each batch offline before training by taking into account the stochasticity in forming batches during training. These pre-calculated kernels can then be used during training of GDKL. This is an advantage of GDKL over the standard NNGP model, which requires computation of the full kernel matrix.  
\end{itemize}

Finally, we would like to highlight two important points. First, although the training of GDKL can be slower compared to DKL, when making predictions the models are equivalent. This is unlike the NNGP model which scales linearly with the number of training points. Second, there are ongoing efforts to scale NNGP models to larger datasets, as evident by recent studies such as \citep{adlam2023kernel}. These advancements in scaling NNGP models could potentially be leveraged in GDKL as well, if needed, to further improve its scalability and applicability to larger datasets.

\textbf{Comparison to baseline methods.}
The GDKL objective consists of two components in the loss function: a predictive distribution term and a KL divergence term. The most computationally intensive factor in calculating both terms is the inversion of the DKL and the NNGP kernel matrices, each with a complexity of $\mathcal{O}((n / 2)^3)$ in the exact case. The division by 2 is due to GDKL partitioning the data into two halves at each iteration, using one half for making predictions on the other half.  
When employing $m$ inducing points, the complexity of the computation involves the inverse of the kernel over the inducing locations and the inverse of the NNGP kernel over half of the examples in the batch, resulting in a complexity of $\mathcal{O}(m^3 + (\mathcal{B} / 2)^3)$, where $\mathcal{B}$ is the batch size. A potential speedup can be achieved by pre-calculating the inverse of the NNGP kernel offline before training, and using it during training. This way, during training the model's computational speed would be similar to standard DKL models.  

To estimate the training time difference between the methods we measured the average time per-iteration in seconds on CIFAR-10 based on $100$ iterations five times. In \tblref{app_tab:calc_time} we present these timing along with the constant time taken for computing the NNGP kernel before training.  According to the data presented in the table, GDKL tends to exhibit slightly slower performance compared to DKL and DUE. However, it's worth noting that the current implementation of the code is not highly optimized, and there are several aspects that can be improved, such as the computation time for the NNGP kernel and the average iteration time.

\begin{table*}[!h]
    \centering
    \caption{Average run time (Sec.) on CIFAR-10 with $\{50, 100, 200, 400, 800\}$ training examples. Results are based on $100$ iterations done $5$ times.}
    \scalebox{.85}{
        \begin{tabular}{l c ccccc}
        \toprule
        && 50 & 100 & 200 & 400 & 800\\
        \midrule
        NNGP (One Time) &&  5.44 $\pm$ 0.32 & 5.51 $\pm$ 0.07 & 5.91 $\pm$ 0.10 & 8.12 $\pm$ 0.23 & 16.70 $\pm$ 0.16\\
        \hline
        DKL &&  0.06 $\pm$ 0.00 & 0.12 $\pm$ 0.00 & 0.22 $\pm$ 0.00 & 0.44 $\pm$ 0.00 & 0.65 $\pm$ 0.00\\
        DLVKL && 0.10 $\pm$ 0.01 & 0.18 $\pm$ 0.00 & 0.29 $\pm$ 0.00 & 0.56 $\pm$ 0.00 & 0.84 $\pm$ 0.00\\
        DUE && 0.08 $\pm$ 0.00 & 0.14 $\pm$ 0.00 & 0.23 $\pm$ 0.00 & 0.45 $\pm$ 0.00 & 0.67 $\pm$ 0.00\\
        GDKL && 0.07 $\pm$ 0.00 & 0.14 $\pm$ 0.00 & 0.24 $\pm$ 0.00 & 0.48 $\pm$ 0.00 & 0.76 $\pm$ 0.04\\
        \bottomrule
        \end{tabular}
    }
    \label{app_tab:calc_time}
    \end{table*}

\section{Additional Experiments}
\label{app_sec:add_exp}

\subsection{Objective Functions Analysis}
\label{app_sec:obj_fun_analysis}
\begin{table*}[!h]
\centering
\caption{Ablation on objective functions - test results on the UCI datasets based on ten random splits.}
\scalebox{.85}{
    \begin{tabular}{l c cccc cccc cccc}
    \toprule
    && \multicolumn{3}{c}{Boston} && \multicolumn{3}{c}{Energy} && 
    \multicolumn{3}{c}{Concrete}\\
    \cmidrule(l){3-5}  \cmidrule(l){7-9} \cmidrule(l){11-13}
    && LL ($\uparrow$) && RMSE ($\downarrow$) && LL ($\uparrow$) && RMSE ($\downarrow$) && LL ($\uparrow$) && RMSE ($\downarrow$)\\
    \midrule
    NNGP && -2.49 $\pm$ 0.16 && 3.19 $\pm$ 0.70 && -1.07 $\pm$ 0.04 && 0.69 $\pm$ 0.02 && -3.15 $\pm$ 0.04 && 4.71 $\pm$ 0.49 \\
    \midrule
    $\ell_{dist}$ && \textbf{-2.51 $\pm$ 0.13} && 3.32 $\pm$ 0.52 && -1.23 $\pm$ 0.04 && 0.72 $\pm$ 0.08 && -3.03 $\pm$ 0.14 && 5.02 $\pm$ 0.58 \\
    $\ell_{pred}$ && -499. $\pm$ 229. && 3.27 $\pm$ 0.92 && -1.78 $\pm$ 1.10 && \textbf{0.28 $\pm$ 0.07} && -6.72 $\pm$ 1.52 && \textbf{4.21 $\pm$ 0.90} \\
    \midrule
    GDKL && \textbf{-2.49 $\pm$ 0.20} && \textbf{3.03 $\pm$ 0.54} && \textbf{-0.50 $\pm$ 0.07} && \textbf{0.32 $\pm$ 0.06} && \textbf{-2.93 $\pm$ 0.10} && 4.58 $\pm$ 0.57 \\
    \bottomrule
    \end{tabular}
}
\label{app_tab:objectives}
\end{table*}

Here we study the effect of using the GDKL objective vs $\ell_{dist}$ and $\ell_{pred}$ which were presented in \Secref{sec:GDKL}. We evaluated all methods on the UCI datasets Boston, Concrete, and Energy according to the setup described in \Secref{sec:small_sized_data}. The results are presented in \tblref{app_tab:objectives}. The table shows that the results are in agreement with our intuition. First, $\ell_{dist}$ behavior is similar to that of the NNGP, the model that it tries to distill. Second, $\ell_{pred}$ clearly overfits as indicated by the log-likelihood values, yet according to the RMSE it is able to maintain a good mean prediction. And lastly, our proposed approach balances well between these two edges, it presents the best results on both metrics in almost all cases.

\subsection{Comparison to a standard Neural Network}
\label{sec:comparison_to_nn}

\begin{table*}[!h]
\centering
\caption{Comparison to a standard NN - test results on CIFAR-10 with $\{50, 100, 200, 400, 800\}$ training examples.}
\scalebox{.65}{
    \begin{tabular}{l c ccccc c ccccc}
    \toprule
    && \multicolumn{5}{c} {Log-Likelihood} && \multicolumn{5}{c}{Accuracy} \\
    \cmidrule(l){3-7}  \cmidrule(l){9-13}
    && 50 & 100 & 200 & 400 & 800 && 50 & 100 & 200 & 400 & 800\\
    \midrule
    NN && -4.55 $\pm$ 0.17 & -4.63 $\pm$ 0.27 & -4.20 $\pm$ 0.26 & -3.76 $\pm$ 0.34 & -3.28 $\pm$ 0.21 &&  19.44 $\pm$ 1.17 & 20.94 $\pm$ 1.90 & 27.82 $\pm$ 1.67 & 34.96 $\pm$ 2.48 & 42.81 $\pm$ 2.06\\
    GDKL (Ours) && \textbf{-2.29 $\pm$ 0.01} & \textbf{-2.08 $\pm$ 0.03} & \textbf{-1.83 $\pm$ 0.01} & \textbf{-1.66 $\pm$ 0.01} & \textbf{-1.49 $\pm$ 0.01} &&  19.35 $\pm$ 0.65 & \textbf{26.53 $\pm$ 0.98} & \textbf{36.45 $\pm$ 0.72} & \textbf{42.97 $\pm$ 0.71} & \textbf{49.75 $\pm$ 0.94}\\
    \bottomrule
    \end{tabular}
}
\label{app_tab:comparison_nn}
\end{table*}
Here we compare GDKL to a standard NN on the CIFAR-10 dataset under the setup outlined in \Secref{sec:high_dim_data}. We present the test results when varying the number of training examples from $50$ to $800$ based on ten random seeds in \tblref{app_tab:comparison_nn}. From the table, GDKL demonstrates superior performance compared to a standard NN in terms of both log-likelihood and accuracy in almost all cases. Furthermore, when cross referencing these results with those in \Figref{fig:high_dim_data}, in general, GP-based methods outperform standard NNs in these experiments conducted under low-data regime conditions.

\subsection{Reliability diagrams}
\label{app_sec:reliability_diagrams}
\begin{figure*}[!b]
    \centering
    \includegraphics[width=1.\textwidth]{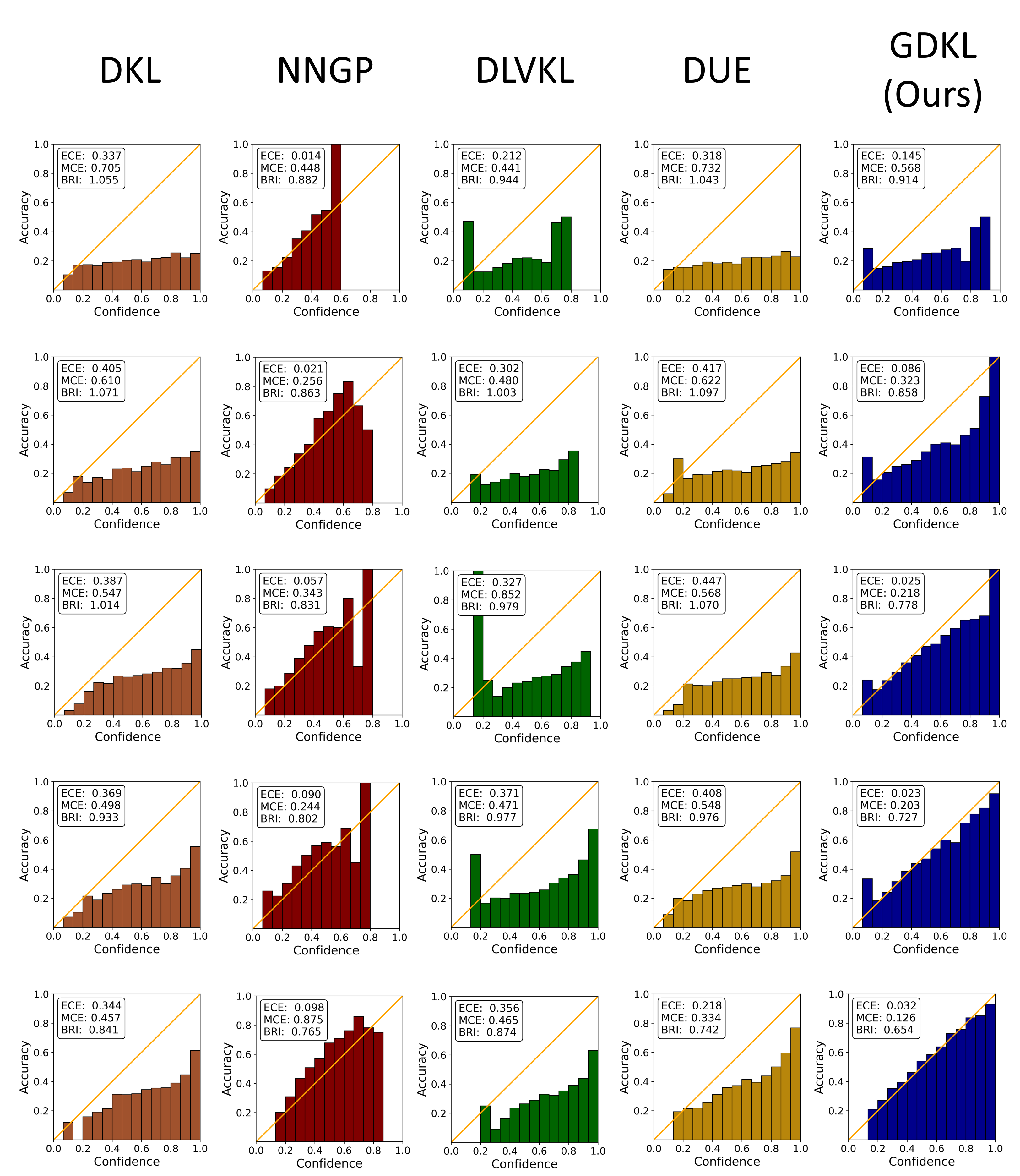}
    \caption{Reliability diagrams on CIFAR-10 test set for the experiments in \Secref{sec:high_dim_data} with training examples ranging from $50$ (top row) to $800$ (bottom row) examples.}
    \label{app_fig:calibration}
\end{figure*}

Here we quantify the confidence through calibration for GDKL and baseline methods on the CIFAR-10 dataset in the setting described in \Secref{sec:high_dim_data}. We use reliability diagrams and the following metrics \citep{brier1950verification, guo2017calibration}: (1) Expected Calibration Error (ECE), which measures the weighted average distance between the classifier confidence and accuracy; (2) Maximum Calibration Error (MCE) which measures the maximum distance between the classifier confidence and accuracy; and (3) Brier score (BRI) which measures the average squared error between the prediction probabilities and the actual labels. \Figref{app_fig:calibration} shows that GDKL is best calibrated across all metrics in all cases when $n \geq 200$, and on smaller dataset sizes only the NNGP model is better. We note that temperature scaling can improve calibration, yet finding the right temperature requires having an additional validation set.

\subsection{Full Results}
In this section, we provide full numerical results for the experiments described in Sections \ref{sec:small_sized_data} and \ref{sec:high_dim_data}. 

\begin{table*}[!h]
\centering
\caption{Train results on small UCI datasets based on ten random splits.}
\scalebox{.85}{
    \begin{tabular}{l c cccc cccc cccc}
    \toprule
    && \multicolumn{3}{c}{Boston} && \multicolumn{3}{c}{Energy} && 
    \multicolumn{3}{c}{Concrete}\\
    \cmidrule(l){3-5}  \cmidrule(l){7-9} \cmidrule(l){11-13}
    && LL ($\uparrow$) && RMSE ($\downarrow$) && LL ($\uparrow$) && RMSE ($\downarrow$) && LL ($\uparrow$) && RMSE ($\downarrow$)\\
    \midrule
    DKL && 1.45 $\pm$ 0.01 && 0.00 $\pm$ 0.00 && 1.36 $\pm$ 0.00 && 0.00 $\pm$ 0.00 && -1.89 $\pm$ 0.47 && 1.28 $\pm$ 0.03 \\
    NNGP && -2.07 $\pm$ 0.05 && 1.66 $\pm$ 0.09 && 1.00 $\pm$ 0.01 && 0.01 $\pm$ 0.00 && -2.96 $\pm$ 0.01 && 2.56 $\pm$ 0.07 \\
    GP-RBF && -1.85 $\pm$ 0.09 && 1.31 $\pm$ 0.16 && -0.18 $\pm$ 0.03 && 0.27 $\pm$ 0.01 && -2.15 $\pm$ 0.05 && 1.70 $\pm$ 0.01 \\
    \midrule
    GDKL && -1.91 $\pm$ 0.04 && 1.04 $\pm$ 0.06 && -0.08 $\pm$ 0.04 && 0.03 $\pm$ 0.00 && -2.64 $\pm$ 0.02 && 2.71 $\pm$ 0.07 \\
    \bottomrule
    \end{tabular}
}
\label{app_tab:fig2_train}
\end{table*}

\begin{table*}[!h]
\centering
\caption{Test results on small UCI datasets based on ten random splits.}
\scalebox{.85}{
    \begin{tabular}{l c cccc cccc cccc}
    \toprule
    && \multicolumn{3}{c}{Boston} && \multicolumn{3}{c}{Energy} && 
    \multicolumn{3}{c}{Concrete}\\
    \cmidrule(l){3-5}  \cmidrule(l){7-9} \cmidrule(l){11-13}
    && LL ($\uparrow$) && RMSE ($\downarrow$) && LL ($\uparrow$) && RMSE ($\downarrow$) && LL ($\uparrow$) && RMSE ($\downarrow$)\\
    \midrule
    DKL && -553. $\pm$ 265. && 3.12 $\pm$ 0.71 && -3.59 $\pm$ 2.09 && 0.32 $\pm$ 0.07 && -3.89 $\pm$ 0.66 && 3.96 $\pm$ 0.63 \\
    NNGP && -2.49 $\pm$ 0.16 && 3.19 $\pm$ 0.70 && -1.07 $\pm$ 0.04 && 0.69 $\pm$ 0.02 && -3.15 $\pm$ 0.04 && 4.71 $\pm$ 0.49 \\
    GP-RBF && -2.39 $\pm$ 0.21 && 2.82 $\pm$ 0.64 && -0.51 $\pm$ 0.13 && 0.40 $\pm$ 0.05 && -2.90 $\pm$ 0.23 && 5.59 $\pm$ 0.80 \\
    \midrule
    GDKL && -2.47 $\pm$ 0.14 && 3.03 $\pm$ 0.49 && -0.31 $\pm$ 0.06 && 0.28 $\pm$ 0.03 && -2.92 $\pm$ 0.09 && 4.58 $\pm$ 0.60 \\
    \bottomrule
    \end{tabular}
}
\label{app_tab:fig2_test}
\end{table*}

\begin{table*}[!h]
\centering
\caption{Test results on Buzz, CTSlice, and CIFAR-10 - 50 examples.}
\scalebox{.85}{
    \begin{tabular}{l c cccc cccc cccc}
    \toprule
    && \multicolumn{3}{c}{Buzz} && \multicolumn{3}{c}{CTSlice} && 
    \multicolumn{3}{c}{CIFAR-10}\\
    \cmidrule(l){3-5}  \cmidrule(l){7-9} \cmidrule(l){11-13}
    && LL ($\uparrow$) && RMSE ($\downarrow$) && LL ($\uparrow$) && RMSE ($\downarrow$) && LL ($\uparrow$) && Acc. ($\uparrow$)\\
    \midrule
    DKL && -421. $\pm$ 208. && 1.61 $\pm$ 0.26 && -283. $\pm$ 577. && 20.07 $\pm$ 3.54 && - 2.61 $\pm$ 0.05 && 20.26 $\pm$ 0.97 \\
    NNGP && -1.29 $\pm$ 0.04 && 1.06 $\pm$ 0.13 && -3.94 $\pm$ 0.05 && 12.35 $\pm$ 0.93 && - 2.23 $\pm$ 0.00 && 18.87 $\pm$ 0.16 \\
    GP-RBF && -1.61 $\pm$ 0.27 && 1.30 $\pm$ 0.16 && -4.53 $\pm$ 0.00 && 22.34 $\pm$ 0.15 && -- $\pm$ -- && -- $\pm$ -- \\
    DLVKL && -- $\pm$ -- && -- $\pm$ -- && -- $\pm$ -- && -- $\pm$ -- && - 2.42 $\pm$ 0.04 && 18.79 $\pm$ 0.86 \\
    DUE && -- $\pm$ -- && -- $\pm$ -- && -- $\pm$ -- && -- $\pm$ -- && - 2.64 $\pm$ 0.07 && 20.18 $\pm$ 0.85 \\
    \midrule
    GDKL && -1.23 $\pm$ 0.09 && 0.86 $\pm$ 0.08 && -3.92 $\pm$ 0.11 && 11.73 $\pm$ 1.10 && - 2.29 $\pm$ 0.02 && 19.35 $\pm$ 0.65 \\
    \bottomrule
    \end{tabular}
}
\label{app_tab:fig3_50}
\end{table*}

\begin{table*}[!h]
\centering
\caption{Test results on Buzz, CTSlice, and CIFAR-10 - 100 examples.}
\scalebox{.85}{
    \begin{tabular}{l c cccc cccc cccc}
    \toprule
    && \multicolumn{3}{c}{Buzz} && \multicolumn{3}{c}{CTSlice} && 
    \multicolumn{3}{c}{CIFAR-10}\\
    \cmidrule(l){3-5}  \cmidrule(l){7-9} \cmidrule(l){11-13}
    && LL ($\uparrow$) && RMSE ($\downarrow$) && LL ($\uparrow$) && RMSE ($\downarrow$) && LL ($\uparrow$) && Acc. ($\uparrow$)\\
    \midrule
    DKL && -336. $\pm$ 185. && 1.38 $\pm$ 0.25 && -518. $\pm$ 350. && 13.10 $\pm$ 5.09 && - 2.73 $\pm$ 0.04 && 24.67 $\pm$ 1.26 \\
    NNGP && -1.18 $\pm$ 0.04 && 0.98 $\pm$ 0.14 && -3.75 $\pm$ 0.03 && 10.25 $\pm$ 0.46 && - 2.14 $\pm$ 0.00 && 22.47 $\pm$ 0.17 \\
    GP-RBF && -1.36 $\pm$ 0.27 && 1.10 $\pm$ 0.09 && -4.52 $\pm$ 0.00 && 22.33 $\pm$ 0.14 && -- $\pm$ -- && -- $\pm$ -- \\
    DLVKL && -- $\pm$ -- && -- $\pm$ -- && -- $\pm$ -- && -- $\pm$ -- && - 2.40 $\pm$ 0.07 && 22.81 $\pm$ 1.68 \\
    DUE && -- $\pm$ -- && -- $\pm$ -- && -- $\pm$ -- && -- $\pm$ -- && - 2.77 $\pm$ 0.14 && 24.32 $\pm$ 1.53 \\
    \midrule
    GDKL && -1.14 $\pm$ 0.05 && 0.77 $\pm$ 0.06 && -3.64 $\pm$ 0.08 && 9.05 $\pm$ 0.77 && - 2.08 $\pm$ 0.03 && 26.53 $\pm$ 0.98 \\
    \bottomrule
    \end{tabular}
}
\label{app_tab:fig3_100}
\end{table*}

\begin{table*}[!h]
\centering
\caption{Test results on Buzz, CTSlice, and CIFAR-10 - 200 examples.}
\scalebox{.85}{
    \begin{tabular}{l c cccc cccc cccc}
    \toprule
    && \multicolumn{3}{c}{Buzz} && \multicolumn{3}{c}{CTSlice} && 
    \multicolumn{3}{c}{CIFAR-10}\\
    \cmidrule(l){3-5}  \cmidrule(l){7-9} \cmidrule(l){11-13}
    && LL ($\uparrow$) && RMSE ($\downarrow$) && LL ($\uparrow$) && RMSE ($\downarrow$) && LL ($\uparrow$) && Acc. ($\uparrow$)\\
    \midrule
    DKL && -178. $\pm$ 127. && 1.39 $\pm$ 0.21 && -459. $\pm$ 210. && 8.18 $\pm$ 0.74 && - 2.58 $\pm$ 0.04 && 33.65 $\pm$ 1.24 \\
    NNGP && -1.09 $\pm$ 0.02 && 0.91 $\pm$ 0.10 && -3.58 $\pm$ 0.05 && 8.89 $\pm$ 0.36 && - 2.02 $\pm$ 0.00 && 28.94 $\pm$ 0.13 \\
    GP-RBF && -1.09 $\pm$ 0.08 && 0.90 $\pm$ 0.11 && -4.42 $\pm$ 0.19 && 20.21 $\pm$ 3.41 && -- $\pm$ -- && -- $\pm$ -- \\
    DLVKL && -- $\pm$ -- && -- $\pm$ -- && -- $\pm$ -- && -- $\pm$ -- && - 2.42 $\pm$ 0.09 && 27.59 $\pm$ 1.68 \\
    DUE && -- $\pm$ -- && -- $\pm$ -- && -- $\pm$ -- && -- $\pm$ -- && - 2.56 $\pm$ 0.21 && 33.10 $\pm$ 1.18 \\
    \midrule
    GDKL && -1.08 $\pm$ 0.04 && 0.77 $\pm$ 0.06 && -3.45 $\pm$ 0.07 && 7.73 $\pm$ 0.66 && - 1.83 $\pm$ 0.01 && 36.45 $\pm$ 0.72 \\
    \bottomrule
    \end{tabular}
}
\label{app_tab:fig3_200}
\end{table*}

\begin{table*}[!h]
\centering
\caption{Test results on Buzz, CTSlice, and CIFAR-10 - 400 examples.}
\scalebox{.85}{
    \begin{tabular}{l c cccc cccc cccc}
    \toprule
    && \multicolumn{3}{c}{Buzz} && \multicolumn{3}{c}{CTSlice} && 
    \multicolumn{3}{c}{CIFAR-10}\\
    \cmidrule(l){3-5}  \cmidrule(l){7-9} \cmidrule(l){11-13}
    && LL ($\uparrow$) && RMSE ($\downarrow$) && LL ($\uparrow$) && RMSE ($\downarrow$) && LL ($\uparrow$) && Acc. ($\uparrow$)\\
    \midrule
    DKL && -136. $\pm$ 82.2 && 1.34 $\pm$ 0.35 && -285. $\pm$ 130. && 7.22 $\pm$ 2.23 && - 2.36 $\pm$ 0.10 && 40.96 $\pm$ 0.89 \\
    NNGP && -1.00 $\pm$ 0.01 && 0.86 $\pm$ 0.09 && -3.39 $\pm$ 0.03 && 7.57 $\pm$ 0.24 && - 1.92 $\pm$ 0.00 && 34.39 $\pm$ 0.09 \\
    GP-RBF && -0.97 $\pm$ 0.03 && 0.79 $\pm$ 0.05 && -3.77 $\pm$ 0.29 && 10.84 $\pm$ 0.86 && -- $\pm$ -- && -- $\pm$ -- \\
    DLVKL && -- $\pm$ -- && -- $\pm$ -- && -- $\pm$ -- && -- $\pm$ -- && - 2.43 $\pm$ 0.09 && 33.48 $\pm$ 1.54 \\
    DUE && -- $\pm$ -- && -- $\pm$ -- && -- $\pm$ -- && -- $\pm$ -- && - 2.20 $\pm$ 0.21 && 40.33 $\pm$ 0.76 \\
    \midrule
    GDKL && -1.01 $\pm$ 0.03 && 0.71 $\pm$ 0.04 && -3.20 $\pm$ 0.05 && 5.98 $\pm$ 0.42 && - 1.66 $\pm$ 0.01 && 42.97 $\pm$ 0.71 \\
    \bottomrule
    \end{tabular}
}
\label{app_tab:fig3_400}
\end{table*}

\begin{table*}[!h]
\centering
\caption{Test results on Buzz, CTSlice, and CIFAR-10 - 800 examples.}
\scalebox{.85}{
    \begin{tabular}{l c cccc cccc cccc}
    \toprule
    && \multicolumn{3}{c}{Buzz} && \multicolumn{3}{c}{CTSlice} && 
    \multicolumn{3}{c}{CIFAR-10}\\
    \cmidrule(l){3-5}  \cmidrule(l){7-9} \cmidrule(l){11-13}
    && LL ($\uparrow$) && RMSE ($\downarrow$) && LL ($\uparrow$) && RMSE ($\downarrow$) && LL ($\uparrow$) && Acc. ($\uparrow$)\\
    \midrule
    DKL && -114. $\pm$ 94.1 && 1.09 $\pm$ 0.12 && -246. $\pm$ 110. && 5.70 $\pm$ 1.47 && - 2.13 $\pm$ 0.10 && 48.57 $\pm$ 0.59 \\
    NNGP && -0.94 $\pm$ 0.01 && 0.81 $\pm$ 0.13 && -3.15 $\pm$ 0.02 && 6.25 $\pm$ 0.19 && - 1.80 $\pm$ 0.06 && 38.73 $\pm$ 0.07 \\
    GP-RBF && -0.89 $\pm$ 0.01 && 0.72 $\pm$ 0.01 && -3.14 $\pm$ 0.11 && 7.61 $\pm$ 0.56 && -- $\pm$ -- && -- $\pm$ -- \\
    DLVKL && -- $\pm$ -- && -- $\pm$ -- && -- $\pm$ -- && -- $\pm$ -- && - 2.22 $\pm$ 0.06 && 43.52 $\pm$ 0.75 \\
    DUE && -- $\pm$ -- && -- $\pm$ -- && -- $\pm$ -- && -- $\pm$ -- && - 1.91 $\pm$ 0.08 && 49.22 $\pm$ 0.79 \\
    \midrule
    GDKL && -0.95 $\pm$ 0.02 && 0.67 $\pm$ 0.03 && -2.99 $\pm$ 0.06 && 4.87 $\pm$ 0.45 && - 1.49 $\pm$ 0.02 && 49.75 $\pm$ 0.94 \\
    \bottomrule
    \end{tabular}
}
\label{app_tab:fig3_800}
\end{table*}

\end{document}